\documentclass[runningheads]{llncs}
\usepackage{graphicx}

\usepackage{tikz}
\usepackage{comment}
\usepackage{amsmath,amssymb} 
\usepackage{color}
\usepackage{orcidlink}

\usepackage[accsupp]{axessibility}  

\definecolor{citecolor}{HTML}{0071bc}
\usepackage{colortbl}  
\usepackage[pagebackref,breaklinks,colorlinks,citecolor=citecolor]{hyperref}

\usepackage[capitalize]{cleveref}
\crefname{section}{Sec.}{Secs.}
\Crefname{section}{Section}{Sections}
\Crefname{table}{Table}{Tables}
\crefname{table}{Tab.}{Tabs.}

\usepackage{tabulary,multirow,xspace}
\usepackage{fixmath,mathtools,nicefrac,mmstyle}
\usepackage{cite}
\usepackage{booktabs}
\usepackage{subcaption}
\usepackage{caption}
\usepackage{multirow}
\usepackage[misc]{ifsym} 
\usepackage{float}

\newcommand{\myparagraph}[1]{{\noindent\bf #1}}
\newlength\savewidth\newcommand\shline{\noalign{\global\savewidth\arrayrulewidth
  \global\arrayrulewidth 1pt}\hline\noalign{\global\arrayrulewidth\savewidth}}
\newcommand{\tablestyle}[2]{\setlength{\tabcolsep}{#1}\renewcommand{\arraystretch}{#2}\centering\small}

\newcommand{\bd}[1]{\textbf{#1}}

\newcolumntype{x}[1]{>{\centering\arraybackslash}p{#1pt}}
\newcommand{\deh}[1]{\textcolor{gray}{#1}}
\definecolor{deemph}{gray}{0.6}

\definecolor{ourscolor}{gray}{.9}
\newcommand{\ours}[1]{\cellcolor{ourscolor}{#1}}

\newcommand{\dist}{\mathcal{D}}
\newcommand{\p}{{p}}  
\newcommand{\z}{{z}}  
\newcommand{\pprime}{{p}^{\prime}}  
\newcommand{\zprime}{{z}^{\prime}}  
\newcommand{\gprime}{{g}^{\prime}}  
\newcommand{\hprime}{{h}^{\prime}}  
\newcommand{\sprime}{{s}^{\prime}}  
\newcommand{\tss}[1]{\textsuperscript{#1}}
\newcommand{\things}{\tss{Th}\xspace}
\newcommand{\stuff}{\tss{St}\xspace}

\begin{document}
\pagestyle{headings}
\mainmatter

\def\ECCVSubNumber{2631}  

\title{Dense Siamese Network for \\Dense Unsupervised Learning} 

\titlerunning{Dense Siamese Network}
%
\author{Wenwei Zhang\orcidlink{0000-0002-2748-4514}\inst{1} \and
Jiangmiao Pang\orcidlink{0000-0002-6711-9319}\inst{2} \and\\
Kai Chen\orcidlink{0000-0002-6820-2325}\inst{2,3} \and
Chen Change Loy\orcidlink{0000-0001-5345-1591}\index{Chen Change, Loy}\inst{1\textrm{\Letter}}}
\authorrunning{W. Zhang et al.}
%
\institute{$^1$S-Lab, Nanyang Technological University \\
$^2$Shanghai AI Laboratory \quad $^3$SenseTime Research\\
\email{\{wenwei001, ccloy\}@ntu.edu.sg}\quad
\email{\{pangjiangmiao, chenkai\}@pjlab.org.cn}}
\maketitle
\begin{abstract}
This paper presents Dense Siamese Network (DenseSiam), a simple unsupervised learning framework for dense prediction tasks.
It learns visual representations by maximizing the similarity between two views of one image with two types of consistency, \ie, pixel consistency and region consistency.
Concretely, DenseSiam first maximizes the pixel level spatial consistency according to the exact location correspondence in the overlapped area. 
It also extracts a batch of region embeddings that correspond to some sub-regions in the overlapped area to be contrasted for region consistency. 
In contrast to previous methods that require negative pixel pairs, momentum encoders or heuristic masks, DenseSiam benefits from the simple Siamese network and optimizes the consistency of different granularities. 
It also proves that the simple location correspondence and interacted region embeddings are effective enough to learn the similarity.
We apply DenseSiam on ImageNet and obtain competitive improvements on various downstream tasks.
We also show that only with some extra task-specific losses, the simple framework can directly conduct dense prediction tasks.
On an existing unsupervised semantic segmentation benchmark, it surpasses state-of-the-art segmentation methods by 2.1 mIoU with 28\% training costs.
Code and models are released at \url{https://github.com/ZwwWayne/DenseSiam}.

\end{abstract}

\section{Introduction}\label{sec:Introduction}

\begin{figure}[t]
	\centering
	\begin{minipage}{.44\linewidth}
		\includegraphics[width=0.95\linewidth]{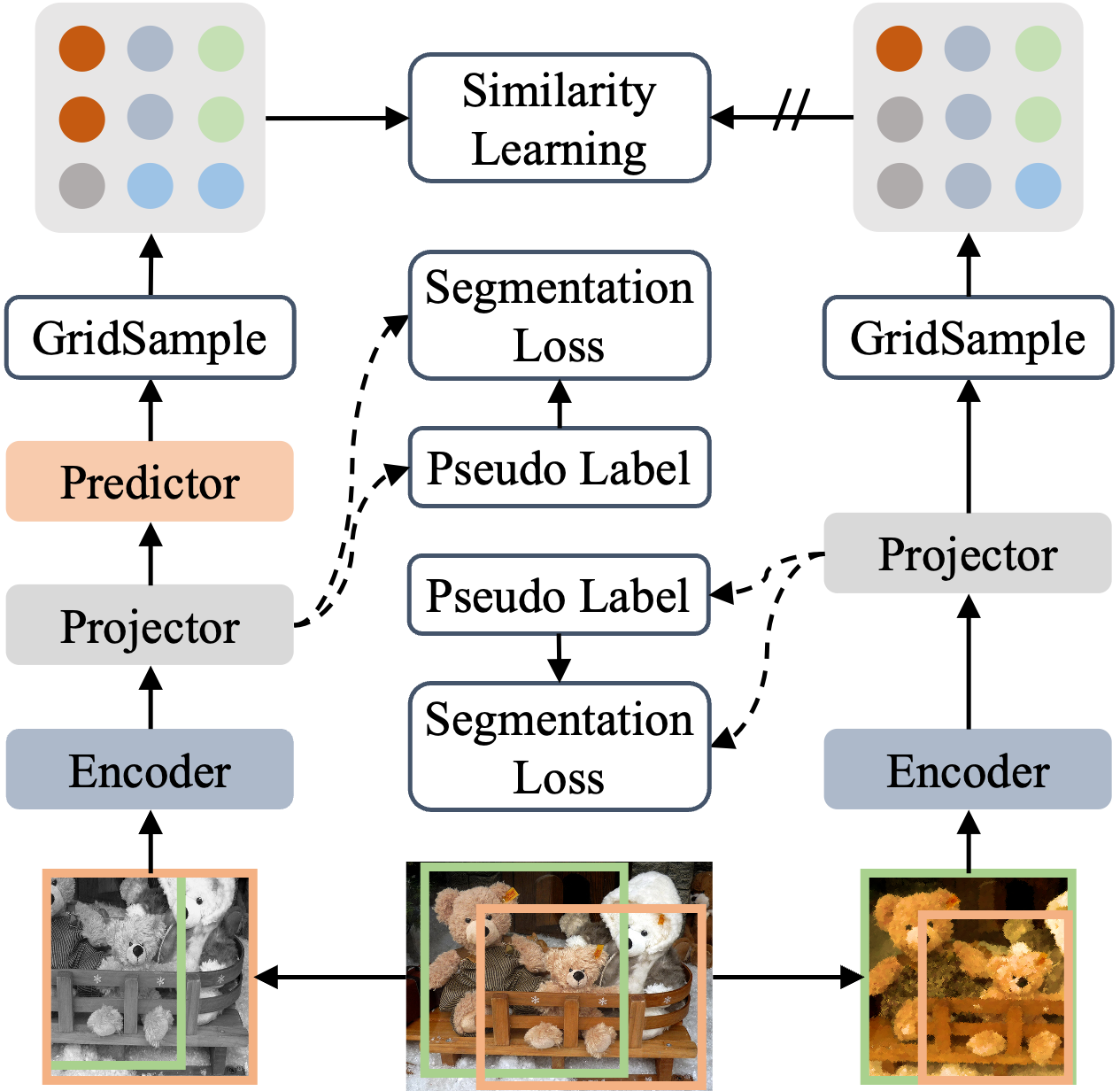}
	\end{minipage}
	\begin{minipage}{.55\linewidth}
		\caption{\small{
		\textbf{Dense Siamese Network} (DenseSiam) unanimously solves unsupervised pre-training and unsupervised semantic segmentation.
		The pair of images are first processed by the same encoder network and a convolutional projector.
		Then a predictor is applied on one side, and a stop-gradient operation on the other side.
		A grid sampling method is used to extract dense predictions in the overlapped area.
		DenseSiam can perform unsupervised semantic segmentation by simply adding a segmentation loss with pseudo labels produced by the projector.
	}}\label{fig:teaser}
	\end{minipage}
\end{figure}

Dense prediction tasks, such as image segmentation and object detection, are fundamental computer vision tasks with many real-world applications.
Beyond conventional supervised learning methods, recent research interests grow in unsupervised learning to train networks from large-scale unlabeled datasets. 
These methods either learn representations as pre-trained weights then fine-tune on downstream tasks~\cite{insloc, simsiam}
or directly learn for specific tasks~\cite{picie}. 

In recent years, unsupervised pre-training has attracted a lot of attention.
Much effort has been geared to learn global representation for image classification~\cite{moco, swav, byol, simclr, simsiam}.
As the global average pooling in these methods discards spatial information, it has been observed~\cite{swav, byol, simclr, insloc} that the learned representations are sub-optimal for dense uses.
Naturally, some attempts conduct similarity learning at pixel-level~\cite{densecl, pixpro, vader} or region-level~\cite{scrl, resim, detcon, detco} and maintain the main structures in global ones to learn representations for dense prediction tasks.

Despite the remarkable progress in that field, the development in unsupervised learning for specific tasks is relatively slow-moving.
For example, solutions in unsupervised semantic segmentation~\cite{iic, picie, autoregressive} rely more on clustering methods (such as k-means) that are also derived from unsupervised image classification~\cite{iic, deepcluster, swav}. 
When reflecting on these similar tasks from the perspective of unsupervised learning, we observe these tasks share the inherent goal of maximizing the similarity of dense predictions (either labels or embeddings) across images but differ in the task-specific training objectives.

In this paper, we propose \textit{Dense Siamese Network} (DenseSiam) to unanimously solve these dense unsupervised learning tasks within a single framework (Fig.~\ref{fig:teaser}).
It learns visual representations by maximizing two types of consistency, \ie, pixel consistency and region consistency, with methods dubbed as \emph{PixSim} and \emph{RegionSim}, respectively. 
The encoder here can be directly fine-tuned for various downstream dense prediction tasks after unsupervised pre-training.
By adding an extra segmentation loss to the projector and regarding the argmaxed prediction of projector as pseudo labels,
the encoder and projector is capable of learning class-wise representations for unsupervised semantic segmentation.

Specifically, PixSim learns to maximize the pixel-level spatial consistency between the grid sampled predictions. 
Its training objective is constrained under the exact location correspondence. 
In addition, the projected feature maps are multiplied with the features from encoder to generate a batch of region embeddings on each image, where each region embedding corresponds to a sub-region in the overlapped area.
RegionSim then conducts contrastive learning between pairs of region embeddings and optimizes them to be consistent.

In contrast to previous unsupervised pre-training methods for dense prediction tasks, DenseSiam benefits from the simple Siamese network~\cite{simsiam} that does not have negative pixel pairs~\cite{densecl, pixpro} and momentum encoders~\cite{pixpro, scrl}. 
Uniquely, DenseSiam optimizes the consistency of different granularities.
The optimization for each granularity is simple as it neither requires heuristic masks~\cite{detcon} nor manual regions crops~\cite{resim, scrl, detco}.

Extensive experiments show that DenseSiam is capable of learning strong representation for dense prediction tasks.
DenseSiam obtains nontrivial 0.4 AP$^\text{mask}$, 0.7 mIoU, and 1.7 AP improvement in comparison with SimSiam when transferring its representation on COCO instance segmentation, Cityscapes semantic segmentation, and PASCAL VOC detection, respectively.
For unsupervised semantic segmentation, DenseSiam makes the first attempt to discard clustering~\cite{picie, iic} while surpassing previous state-of-the-art method~\cite{picie} by \bd{2.1} mIoU with only $\sim$28\% of the original training costs.

\section{Related Work}

\noindent\textbf{Siamese Network.}
Siamese network~\cite{bromley1993signature} was proposed for comparing entities.
They have many applications including object tracking~\cite{siamfc}, face verification~\cite{facenet, deepface}, and one-shot recognition\cite{koch2015siamese}.
In conventional use cases, Siamese Network takes different images as input and outputs either a global embedding of each image for comparison~\cite{facenet, deepface, bromley1993signature, koch2015siamese}
or outputs feature map of each image for cross-correlation~\cite{siamfc}.
DenseSiam uses the Siamese architecture to output pixel embeddings and maximizes similarity between embeddings of the same pixel from two views of an image to pre-train dense representations with strong transferability in dense prediction tasks.

\begin{table}[t]
    \centering
    \small
    \caption{\textbf{Comparison of unsupervised dense representation learning methods}.
    }\label{tab:related_work}
        \tablestyle{2pt}{1.1}
        \scalebox{0.75}{\begin{tabular}{l|c|c|c|c|c}
        \shline
        Method &  Base & Pixel & Region & Extra Components & Correspondence \\ \shline
        DenseCL~\cite{densecl}& MoCo v2~\cite{mocov2}& \checkmark&$\times$&$\times$& feature similarity\\
        PixPro~\cite{pixpro}& BYOL~\cite{byol} & \checkmark &$\times$ &$\times$& coordinate distance\\
        VaDER~\cite{vader} & MoCo v2 & \checkmark &$\times$&$\times$& location \\\hline
        ReSim~\cite{resim} & MoCo v2 &$\times$& \checkmark&$\times$ & dense region crops \\
        SCRL~\cite{scrl} & BYOL~\cite{byol}&$\times$& \checkmark &$\times$& sampled region crops \\
        DetCo~\cite{detco} & MoCo v2 &$\times$& \checkmark & $\times$& image patches \\
        SoCo~\cite{soco} & BYOL&$\times$& \checkmark &selective search\cite{selective_search}& sampled region crops \\
        DetCon~\cite{detcon} & SimCLR~\cite{simclr}&$\times$&\checkmark&FH masks~\cite{FH}& heuristic masks \\
        CAST~\cite{CAST} & MoCo v2 &$\times$&\checkmark&DeepUSPS~\cite{deepusps}&saliency map \\\hline
        DenseSiam & SimSiam~\cite{simsiam} & \checkmark&\checkmark&$\times$& location + region embeddings \\ \shline
        \end{tabular}}
  \end{table}

\noindent\textbf{Unsupervised Representation Learning.}
Representative unsupervised representation learning methods include contrastive learning~\cite{examplar_cnn, instance_discrimination, amdim, simclr, moco, mocov2} and clustering-based methods~\cite{deepcluster, swav, scan, online_clustering}.
Notably, Siamese network has become a common structure in recent attempts~\cite{byol, swav, simclr}, despite their different motivations and solutions to avoid the feature `collapsing' problem.
SimSiam~\cite{simsiam} studies the minimum core architecture in these methods~\cite{byol, swav, simclr} and shows that a simple Siamese network with stopping gradient can avoid feature `collapsing' and yield strong representations.

Given the different natures of image-level representation learning and dense prediction tasks,
more recent attempts~\cite{resim, detco, scrl, insloc, updetr, CAST, detcon, densecl, pixpro, vader, orl, soco} pre-train dense representations specially designed for dense prediction tasks.
Most of these methods conduct similarity learning with pixels~\cite{vader, densecl, pixpro},
manually cropped patches of image or features~\cite{resim, detco, scrl, soco, orl}, and regional features segmented by saliency map~\cite{CAST} or segmentation masks~\cite{detcon} obtained in unsupervised manners.
These methods still need a momentum encoder~\cite{scrl, detco, densecl, pixpro, vader} or negative pixel samples~\cite{densecl, pixpro, vader},
although they~\cite{byol, simclr} have been proven unnecessary~\cite{simsiam} in image-level representation learning.

DenseSiam only needs a Siamese network with a stop gradient operation to obtain strong dense representations, without momentum encoder~\cite{scrl, detco, densecl, pixpro, vader}
nor negative pixel pairs~\cite{densecl, pixpro, vader} (Table~\ref{tab:related_work}).
It conducts contrastive learning among regional features inside an image emerged via pixel similarity learning.
Thanks for the unique design, DenseSiam performs region-level similarity learning without saliency maps~\cite{CAST}, region crops~\cite{scrl, resim, detco}, nor heuristic masks~\cite{detcon}.

\noindent\textbf{Unsupervised Semantic Segmentation.}
Unsupervised semantic segmentation aims to predict labels for each pixel without annotations.
There are a few attempts that introduce heuristic masks and conduct similarity learning between segments (regions)~\cite{maskcontrast, segsort} with object-centric images.
Most methods focus on natural scene images and exploit the assumption that semantic information should be invariant to photometric transformations and equivariant to geometric transformations no matter how the model predicts the labels,
which is inherently consistent with the goal of unsupervised representation learning.
However, these methods still rely on clustering~\cite{picie, iic, autoregressive} to predict the per-pixel labels and maximize the consistency of cluster assignments in different views of the image,
which is cumbersome and difficult to be used for large-scale data.

DenseSiam conducts unsupervised semantic segmentation by adding a class-balanced cross entropy loss without clustering, significantly reduces the training costs and makes it scalable for large-scale data.
RegionSim further boosts the segmentation accuracy by maximizing the consistency between regional features.

\section{Dense Siamese Network}\label{sec:methods}

\begin{figure}[t]
	\centering\includegraphics[width=0.98\linewidth]{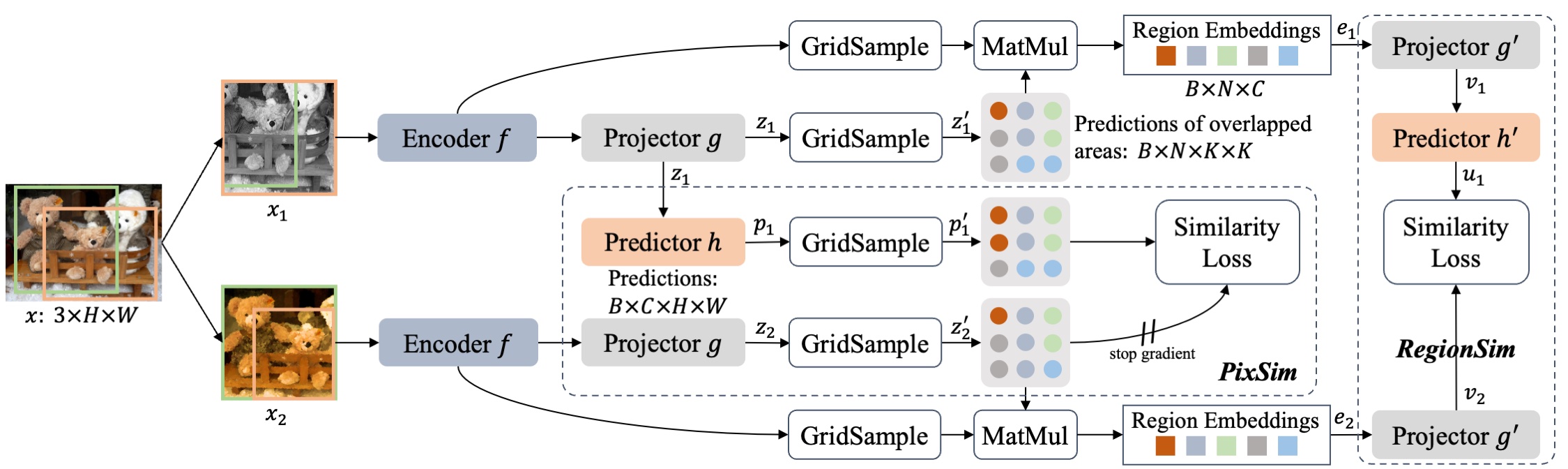}
	\caption{\small{
	  \textbf{Pipeline of DenseSiam.}
	  DenseSiam takes two randomly augmented views of an image $x$ as inputs.
	  The two views are processed by an encoder network $f$ without global average pooling (GAP), followed by a projector network $g$.
	  The predictor network $h$ transforms the dense prediction of the projector of one view.
	  Then GridSample module samples the same pixels inside the intersected regions of the two views and interpolates their feature grids.
	  The similarity between feature grids from two views are maximized by PixSim.
		  Then DenseSiam multiplies the features of encoder and the dense predictions by projector $g$ in the overlapped area to obtain region embeddings of each image.
		  These region embeddings are processed by a new projector $\gprime$ and new predictor $\hprime$ for region level similarity learning.
	}}\label{fig:pipeline}
  \end{figure}

DenseSiam, as shown in Fig.~\ref{fig:pipeline}, is a generic unsupervised learning framework for dense prediction tasks. 
The framework is inspired by SimSiam~\cite{simsiam} but differs in its formulation tailored for learning dense representation (Sec.~\ref{sec:siamese}). 
In particular, it conducts dense similarity learning by PixSim (Sec.~\ref{sec:pixsim}), which aims at maximizing pixel-level spatial consistency between grid sampled predictions. 
Based on the region embeddings derived from per-pixel predictions inferred in PixSim, DenseSiam further performs region-level contrastive learning through RegionSim (Sec.~\ref{sec:regionsim}).

\subsection{Siamese Dense Prediction}\label{sec:siamese}

As shown in Fig.~\ref{fig:pipeline}, DenseSiam takes two randomly augmented views $x_1$ and $x_2$ of an image $x$ as inputs.
The two views are processed by an encoder network $f$ and a projector network $g$.
The encoder network can either be a backbone network like ResNet-50~\cite{resnet} or a combination of networks such as ResNet-50 with FPN~\cite{fpn}.
As the encoder network $f$ does not use global average pooling (GAP), 
the output of $f$ is a \emph{dense feature map} and is later processed by the projector $g$, which consists of three $1\times1$ convolutional layers followed by Batch Normalization (BN) and ReLU activation.
Following the design in SimSiam~\cite{simsiam}, the last BN layer in $g$ does not use learnable affine transformation~\cite{BN}.
The projector $g$ projects the per-pixel embeddings to either a feature space for representation learning or to a labels space for segmentation.
Given the dense prediction from $g$, noted as $\z_1\!\triangleq\!g(f(x_1))$, the predictor network $h$ transforms the output of one view and matches it to another view~\cite{simsiam}.
Its output is denoted as $\p_1\!\triangleq\!h(g(f(x_1)))$.

The encoder and projector essentially form a Siamese architecture, which outputs two dense predictions ($\z_1$ and $\z_2$) for two views of an image, respectively.
DenseSiam conducts similarity learning of different granularities using the Siamese dense predictions with the assistance of predictor~\cite{simsiam, byol}.
After unsupervised pre-training, only the encoder network is fine-tuned for downstream dense prediction tasks, where the projector is only used during pre-training to improve the representation quality~\cite{simclr}.
When directly learning the dense prediction tasks, the encoder and projector are combined to tackle the task.

\subsection{PixSim}\label{sec:pixsim}

\begin{figure}[t]
	\centering
	\begin{minipage}[]{.50\linewidth}
		\includegraphics[width=\linewidth]{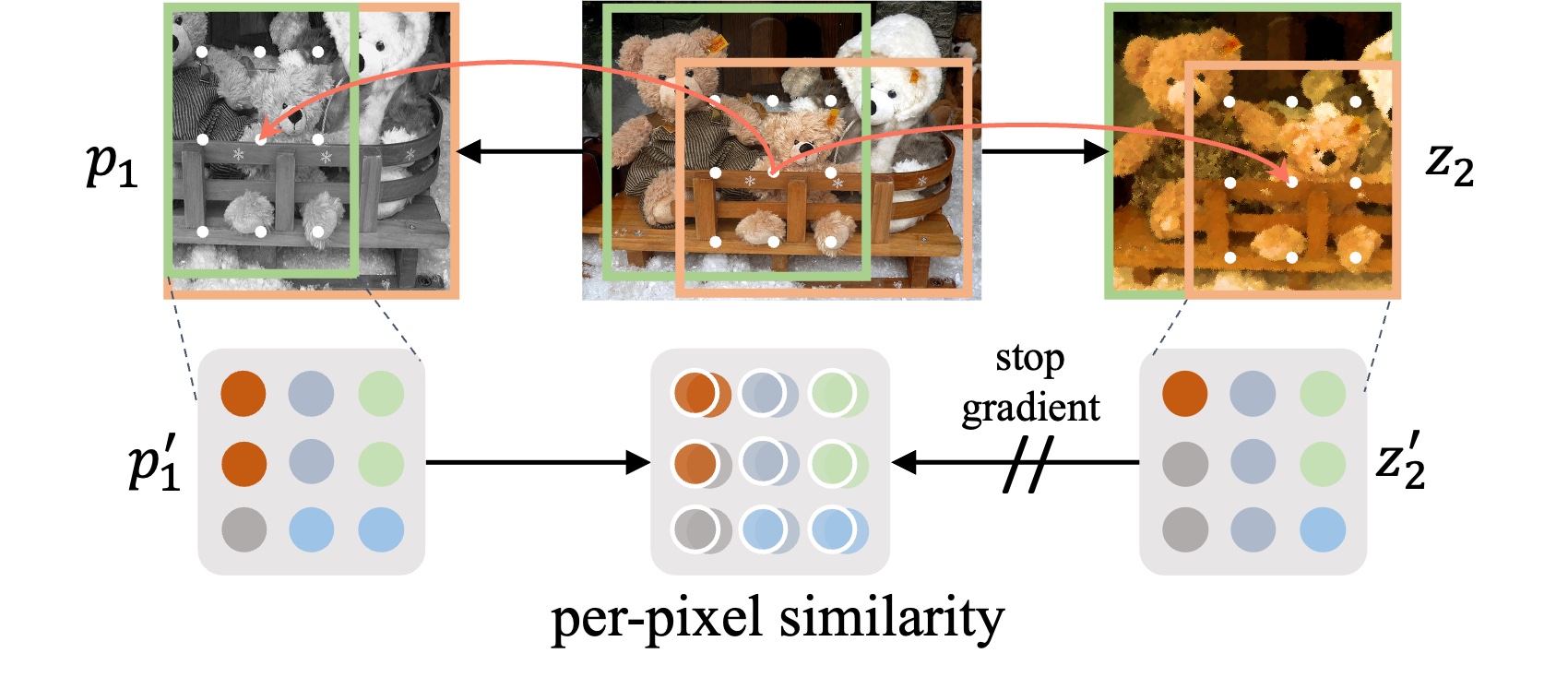}
	\end{minipage}
	\begin{minipage}[]{.48\linewidth}
		\caption{\small{\textbf{Grid Sample module when grid size $K{=}3$.}
		Pixels at the same location in the original image should have similar predictions when they are transformed into two views.
		The Grid Sample module samples these pixels and interpolates their predictions in two views.
		The predictions of one view are learned to match those in another view.
		}}\label{fig:gridsample}
	\end{minipage}
\end{figure}

\myparagraph{Dense Similarity Learning.}
We formulate the dense similarity learning to maximize the similarity of dense predictions inside the overlapping regions between $\p_1$ and $\z_2$.
Specifically, given the relative coordinates of the intersected region in $x_1$ and $x_2$,
we uniformly sample $K\times K$ point grids inside the intersected region from both views as shown in Fig.~\ref{fig:gridsample}.
These points in two views have exactly the same coordinates when they are mapped to the original image $x$;
thus, their predictions should be consistent to those in another view.
Assuming the sampled feature grids are $\zprime_1\!\triangleq\!\texttt{gridsample}(\z_1)$ and $\pprime_1\!\triangleq\!\texttt{gridsample}(\p_1)$, PixSim
minimizes the distance of the feature grids by a symmetrical loss~\cite{byol, simsiam}:
\newcommand{\lnorm}[1]{\frac{#1}{\left\lVert{#1}\right\rVert _2}}
\newcommand{\lnormv}[1]{{#1}/{\left\lVert{#1}\right\rVert _2}}
\newcommand{\logsoftmax}[1]{\log\texttt{softmax}({#1})}
\newcommand{\softmax}[1]{\texttt{softmax}({#1})}
\begin{equation}
\begin{aligned}
\mathcal{L}_{dense}&{=}\frac{1}{2}\dist(\pprime_1, \texttt{stopgrad}(\zprime_2)) \\
&{+}\frac{1}{2}\dist(\pprime_2, \texttt{stopgrad}(\zprime_1)),
\label{eq:loss_sym_stopgrad}
\end{aligned}
\end{equation}
where $\texttt{stopgrad}$ is the stop-gradient operation to prevent feature `collapsing'~\cite{simsiam} and $\dist$ is a distance function that can have many forms~\cite{simsiam, swav}.
Two representative distance functions are negative cosine similarity~\cite{simsiam, byol}
\begin{equation}
\dist(\pprime_1, \zprime_2) {=} - \lnorm{\pprime_1}{\cdot}\lnorm{\zprime_2},
\label{eq:dist_cosine}
\end{equation}
and cross-entropy similarity
\begin{equation}
	\dist(\pprime_1, \zprime_2) {=} - \softmax{\pprime_1}{\cdot}\logsoftmax{\zprime_2}.
	\label{eq:dist_ce}
\end{equation}

\myparagraph{Advantages to Image-Level Similarity Learning.}
Previous formulation of image-level representation learning faces two issues when transferring their representations for dense prediction tasks.
First, it learns global embeddings for each image during similarity learning, where most of the spatial information is discarded during GAP.
Hence, the spatial consistency is not guaranteed in its goal of representation learning,
while dense prediction tasks like semantic segmentation and object detection rely on the spatial information of each pixel.
Second, it is common that $p_1$ and $z_2$ contain different contents (see examples of $x_1$ and $x_2$ in Fig.~\ref{fig:pipeline}) after heavy data augmentations like random cropping and resizing,
but the global embeddings that encode these different contents are forced to be close to each other during training.
This forcefully makes the embeddings of different pixel groups to be close to each other, breaking the regional consistency and is thus undesirable for dense prediction tasks.

PixSim resolves the above-mentioned issues by conducting pixel-wise similarity learning inside the intersected regions of two views,
where the embeddings at identical locations of different views have different values due to augmentations~\cite{simclr} are forced to be similar to each other in training.
Such a process learns dense representations that are invariant to data augmentations,
which are more favorable dense representations for dense prediction tasks.

\subsection{RegionSim}\label{sec:regionsim}

When using cross-entropy with softmax as the training objective,
PixSim implicitly groups similar pixels together into the same regions/segments. The consistency of such regions can be further maximized by the proposed RegionSim to learn better dense representation. 
The connection between PixSim and RegionSim is seamless - region-level similarity learning can be achieved without heuristic masks~\cite{detcon}, saliency map~\cite{CAST}, or manually cropping multiple regional features~\cite{resim, scrl, detco} as in previous studies.  

\myparagraph{Region Embedding Generation.}
As shown in Fig.~\ref{fig:pipeline}, DenseSiam first obtains the feature grids of the intersected regions of each view by grid sampling from the features produced by the encoder $f$.
This restricts RegionSim inside the intersected regions to ensure that the region embeddings to encode the similar contents.
For simplicity, the region embeddings $e_{1} \in \mathcal{R}^{N\times C}$ are obtained by the summation of multiplication between the masks and features as
\begin{equation}
	e_{1} {=} \sum_{i,j}^{K\times K} \zprime_1[i,j] \cdot  \texttt{gridsample}(f(x_1))[i,j],
	\label{eq:mask_embedding}
\end{equation}
where $N$ and $C$ represent the number of sub-regions and the number of feature channels, respectively.
The process yields an embedding for each segment of each pseudo category, which will then be used for contrastive learning to increase the consistency between these region embeddings.

\myparagraph{Region Similarity Learning.}
After obtaining the region embeddings, RegionSim transforms them by a projector network $\gprime$, which is a three-layer MLP head~\cite{byol, simclr, simsiam}.
The output of $\gprime$ is then transformed by $\hprime$ for contrasting or matching with another view.
RegionSim assumes each region embedding of a sub-region to be consistent with the embedding of the sub-region in another view.
Therefore, Eq.~\ref{eq:loss_sym_stopgrad} can be directly applied to these region embeddings to conduct region-level similarity learning.
Meanwhile, we also enforce the region embedding to have low similarities with the embeddings of other region embeddings to make the feature space more compact.
Consequently, by denoting the two outputs as $u_1\!\triangleq\!\hprime(\gprime(e_1))$ and $v_2\!\triangleq\!\gprime(e_2)$, RegionSim can also minimize the symmetrized loss function
\begin{equation}
\begin{aligned}
	\mathcal{L}_{region}&{=}\frac{1}{2}\mathcal{L}_c(u_1, v_2){+}\frac{1}{2}\mathcal{L}_c(u_2, v_1).
	\label{eq:loss_sym_contrast}
\end{aligned}
\end{equation}
The contrastive loss function~\cite{infonce} $\mathcal{L}_c$ can be written as
\begin{equation}
	\mathcal{L}_c(u_1, v_2){=}-\sum_{s}^{N}\log \frac{\exp(u_1^{s} \cdot v_2^{s})}{\sum_{\sprime}^{N}\exp(u_1^{s} \cdot v_2^{\sprime})}.
	\label{eq:dist_contrast}
\end{equation}

\subsection{Learning Objective}\label{subsec:loss}

When conducting unsupervised representation learning,
DenseSiam also maintains a global branch to conduct image-level similarity learning to enhance the global consistency of representations, besides PixSim and RegionSim shown in Fig.~\ref{fig:pipeline}.
The architecture of the global branch remains the same as SimSiam~\cite{simsiam}.
The numbers of channels in the projectors and predictors of PixSim and RegionSim are set to 512 for efficiency.
Denoting the loss of the global branch as $\mathcal{L}_{sim}$, DenseSiam optimizes the following loss
\begin{equation}
	\mathcal{L}{=}\mathcal{L}_{sim} + \lambda_{1}\mathcal{L}_{dense} + \lambda_{2}\mathcal{L}_{region},
	\label{eq:loss_total}
\end{equation}
where $\lambda_{1}$ and $\lambda_{2}$ are loss weights of $\mathcal{L}_{dense}$ for PixSim and $\mathcal{L}_{region}$ for RegionSim, respectively.
The encoder $f$ is used for fine-tuning in downstream dense prediction tasks after pre-training.

\section{Unsupervised Semantic Segmentation}
\label{sec:uss}

The proposed framework is appealing in that it can be readily extended to address dense prediction tasks such as unsupervised semantic segmentation, by simply adding a task-specific losses and layers, without needing the offline and cumbersome clustering process~\cite{autoregressive, iic, picie}.

Formally, when using cross-entropy similarity (Eq.~\ref{eq:dist_ce}) in PixSim,
the softmax output $\softmax{\z}$ can be regarded as the probabilities of belonging to each of $N$ pseudo-categories.
In such a case, the projector $g$ predicts a label for each pixel, which aligns with the formulation for unsupervised semantic segmentation.

DenseSiam uses ResNet with a simplified FPN~\cite{fpn, picie} as the encoder $g$.
The output channel $N$ of PixSim is modified to match the number of classes in the dataset (e.g., 27 on COCO stuff-thing dataset~\cite{coco, picie}).
Our experiment shows that a direct change in the number of output channels in PixSim without any further modification can already yield reasonable performance without feature `collapsing'.
Following previous method~\cite{picie} that encourages the prediction scores to have a lower entropy (\ie, more like one-hot scores),
we add a cross entropy loss, denoted as $\mathcal{L}_{seg}$, when applying PixSim for unsupervised semantic segmentation with the pseudo labels obtained by $\texttt{argmax}(\z_1)$.

We also observe that the small number of categories undermines the training stability, which is also a common issue in clustering-based methods~\cite{picie, iic}.
To solve this issue, DenseSiam introduces another set of projector and predictor in PixSim to keep a large number of pseudo categories following the over-clustering strategy~\cite{picie, iic}.
This head only conducts similarity learning using Eq.~\ref{eq:loss_sym_stopgrad}, noted as $\mathcal{L}_{aux}$.
Therefore, the overall loss of DenseSiam for unsupervised semantic segmentation is calculated as
\begin{equation}
	\mathcal{L}{=}\lambda_{1}\mathcal{L}_{dense} + \lambda_{2}\mathcal{L}_{region} + \lambda_{3}\mathcal{L}_{seg} + \lambda_{4}\mathcal{L}_{aux}.
	\label{eq:loss_total_seg}
\end{equation}
We use $\lambda_{4}{=}\frac{\log N}{\log N + \log N_{aux}}$ and $\lambda_{1}{=}\frac{\log N_{aux}}{\log N + \log N_{aux}}$
to prevent the auxiliary loss from overwhelming the gradients because it uses a larger number $N_{aux}$ of pseudo categories~\cite{picie}.
Note that RegionSim is only used in training to enhance the region consistency of dense labels predicted by PixSim.
Only the encoder $f$ and projector $g$ are combined to form a segmentation model at inference time.


\section{Experiments}\label{sec:Experiments:representation}

\subsection{Experimental Settings}\label{sec:exps:settings}
\myparagraph{Datasets.}
For unsupervised pre-training,
we compare with other methods on ImageNet-1k~\cite{ILSVRC15} (IN1k) dataset and conduct ablation studies on COCO~\cite{coco} dataset as it is smaller.
COCO and IN1k are two large-scale datasets that contain $\sim$118K and $\sim$1.28 million images, respectively.
Though having more images, IN1k is highly curated and it mainly consists of object-centric images. Thus, the data  is usually used for image classification.
In contrast, COCO contains more diverse scenes in the real world and it has more objects ($\sim$7 objects \vs $\sim$1 in IN1k) in one image. Hence, it is mainly used for dense prediction tasks like object detection, semantic, instance, and panoptic segmentation.

For unsupervised semantic segmentation,
the model is trained and evaluated on curated subsets~\cite{iic,picie} of COCO \texttt{train2017} split and \texttt{val2017} split, respectively.
The training and validation set contain 49,629 images and 2,175 images, respectively.
The semantic segmentation annotations include 80 thing categories and 91 stuff categories~\cite{caesar2018coco}.
We follow previous methods~\cite{picie, iic} to merge these categories to form 27 (15 `stuff' and 12 `things') categories. 

\myparagraph{Training Setup.}
We closely follow the pre-training settings of SimSiam~\cite{simsiam} when conducting unsupervised pre-training experiments of DenseSiam.
Specifically, we use a learning rate of $lr{\times}$BatchSize${/}256$ following the linear scaling strategy~\cite{in_1hour}, with a base $lr\!=\!0.05$.
The batch size is 512 by default.
We use the cosine decay learning rate schedule~\cite{SGDR} and SGD optimizer, where the weight decay is $0.0001$ and the SGD momentum is $0.9$.
We pre-train DenseSiam for 800 epochs on COCO and for 200 epochs on IN1k.
ResNet-50~\cite{resnet} is used as default.

For unsupervised semantic segmentation, for a fair comparison, we follow PiCIE~\cite{picie} to adopt the backbone pre-trained on IN1k in a supervised manner.
The backbone is then fine-tuned with DenseSiam for unsupervised semantic segmentation using the same base learning rate, weight decay, and SGD momentum
as those used in unsupervised representation learning. The batch size is 256 by default.
We empirically find the constant learning schedule works better during training. The model is trained for 10 epochs.
ResNet-18~\cite{resnet} is used as default for a fair comparison with previous methods~\cite{picie}.

\myparagraph{Evaluation Protocol of Transfer Learning.}
For unsupervised pre-training, we evaluate the transfer learning performance of the pre-trained representations following Wang~\etal~\cite{densecl}.
We select different dense prediction tasks to comprehensively evaluate the transferability of the dense representation, including semantic segmentation, object detection, and instance segmentation.
Challenging and popular dataset with representative algorithms in each task is selected.

When evaluating on semantic segmentation, we fine-tune a FCN~\cite{FCN} model and evaluate it with the Cityscapes dataset~\cite{cityscapes}.
The model is trained on the \texttt{train\_fine} set (2975 images) for 40k iterations and is tested on the \texttt{val} set.
Strictly following the settings in MMSegmentation~\cite{mmseg2020}, we use FCN-D8 with a crop size of 769, a batch size of 16, and synchronized batch normalization.
Results are averaged over five trials.

When evaluating on object detection, we fine-tune a Faster R-CNN~\cite{ren2015faster} with C4-backbone by 24k iterations on VOC 2007 trainval + 2012 train set and is tested in VOC 2007 test set.
Results are reported as an average over five trials.

We also evaluate the representation on object detection and instance segmentation on COCO dataset~\cite{coco}.
We fine-tune a Mask R-CNN with FPN~\cite{fpn} with standard multi-scale 1x schedule~\cite{mmdetection, wu2019detectron2} on COCO \texttt{train2017} split
and evaluate it on COCO \texttt{val2017} split.
We apply synchronized batch normalization in backbone, neck, and RoI heads during training~\cite{scratch, densecl}.

\myparagraph{Evaluation of Unsupervised Semantic Segmentation.}
Since the model is trained without labels, a mapping between the model's label space and the ground truth categories needs be established.
Therefore, we first let the model predicts on each image on the validation set, then we calculate the confusion matrix between the predicted labels and the ground truth classes.
We use linear assignment to build a one-to-one mapping between the predicted labels and ground truth classes by taking the confusion matrix as the assignment cost.
Then we calculate mean IoU over all classes based on the obtained mapping~\cite{picie, iic}.
To more comprehensively understand the model's behavior, we also report mean IoU of stuff and things classes, noted as mIoU\stuff and mIoU\things, respectively.

\subsection{Transfer Learning Results}
The comparison on transfer learning in dense prediction tasks between DenseSiam and previous unsupervised representation learning methods~\cite{moco, byol, simclr, densecl, resim, simsiam} is shown in Table~\ref{tab:transfer}.
The results of scratch, supervised, MoCo v2~\cite{mocov2}, DenseCL~\cite{densecl}, SimCLR~\cite{simclr}, and BYOL~\cite{byol} are reported from DenseCL~\cite{densecl}.
For fair comparison, we fine-tune the model of ReSim-C4~\cite{resim} using the similar setting (Sec.~\ref{sec:exps:settings}).
The model checkpoint is released by the paper authors\footnote{\url{https://github.com/Tete-Xiao/ReSim}}.
We report the results of SimSiam~\cite{simsiam}, DetCo~\cite{detco}, and PixPro~\cite{pixpro} based on our re-implementation.

\begin{table*}[t]
  \centering
  \caption{\textbf{Transfer Learning}.
  All unsupervised methods are either based on 200-epoch pre-training in ImageNet (`IN1k'). 
  \emph{COCO instance segmentation} and \emph{COCO detection}:  Mask R-CNN~\cite{mask_rcnn} (1$\times$ schedule) fine-tuned in COCO \texttt{train2017};
  \emph{Cityscapes}: FCN fine-tuned on Cityscapes dataset~\cite{cityscapes};
  \emph{VOC 07+12 detection}: Faster R-CNN fine-tuned on VOC 2007 trainval + 2012 train, evaluated on VOC 2007 test;
  COCO \texttt{train2017}, evaluated on COCO \texttt{val2017}.
  All Mask R-CNN are with FPN~\cite{fpn}.
  All Faster R-CNN models are with the C4-backbone~\cite{Detectron2018}.
  All VOC and Cityscapes results are averaged over 5 trials
  }\label{tab:transfer}
  \tablestyle{3pt}{1.1}
  \scalebox{0.8}{\begin{tabular}{l |x{28}x{28}x{28} | x{26}x{26}x{26}|c|x{26}x{26}x{26} }
  \shline
  & \multicolumn{3}{c|}{COCO Instance Seg.}
  & \multicolumn{3}{c|}{COCO Detection}
  & Cityscapes & \multicolumn{3}{c}{VOC 07+12 Detection}\\
  Pre-train   
  & AP$^\text{mask}$ & AP$^\text{mask}_\text{50}$ & AP$^\text{mask}_\text{75}$
  & AP & AP$_\text{50}$ & AP$_\text{75}$
  & mIoU
  & AP & AP$_\text{50}$ & AP$_\text{75}$\\
  \shline
  \deh{scratch} & \deh{29.9} & \deh{47.9} & \deh{32.0} & \deh{32.8} & \deh{50.9} & \deh{35.3} & \deh{63.5} & \deh{32.8} & \deh{59.0} & \deh{31.6}   \\
  supervised    & 35.9 & 56.6 & 38.6 & 39.7 & 59.5 & 43.3 & 73.7 & 54.2 & 81.6 & 59.8 \\
  \hline
  BYOL~\cite{byol}        &34.9 &55.3 &37.5 &38.4 &57.9 &41.9 &71.6  &51.9 &81.0 &56.5  \\
  SimCLR~\cite{simclr}    &34.8 &55.2 &37.2 &38.5 &58.0 &42.0 &73.1  &51.5 &79.4 &55.6  \\
  MoCo v2~\cite{mocov2}   &36.1 &56.9 &38.7 &39.8 &59.8 &43.6 &74.5  &57.0 &82.4 &63.6  \\
  SimSiam~\cite{simsiam}  &36.4	&57.4	&38.8 &40.4	&60.4	&44.1 &76.3  &56.7 &82.3 &63.4	\\\hline
  ReSim~\cite{resim}      &36.1 &56.7 &38.8 &40.0 &59.7 &44.3 &76.8  &58.7 &83.1 &66.3  \\
  DetCo~\cite{detco}      &36.4 &57.0 &38.9 &40.1 &60.3 &43.9 &76.5  &57.8 &82.6 &64.2  \\
  DenseCL~\cite{densecl}  &36.4 &57.0 &39.2 &40.3 &59.9 &44.3 &75.7  &58.7 &82.8 &65.2  \\
  PixPro~\cite{pixpro}    &36.6	&57.3	&39.1 &40.5	&60.1	&44.3 &76.3  &\bd{59.5}&\bd{83.4}&\bd{66.9}   \\
  \ours{DenseSiam} &\ours{\bd{36.8}}&\ours{\bd{57.6}}&\ours{\bd{39.8}}
  &\ours{\bd{40.8}} &\ours{\bd{60.7}}&\ours{\bd{44.6}}
  &\ours{\bd{77.0}} &\ours{58.5}	   &\ours{82.9}	&\ours{65.3} \\\shline
  \end{tabular}}
\end{table*}

\myparagraph{COCO Instance Segmentation and Detection.}
As shown in the first two columns of Table~\ref{tab:transfer},
DenseSiam outperforms SimSiam by 0.4 AP$^\text{mask}$ on COCO instance segmentation.
The improvements over SimSiam is on-par with that of DenseCL and DetCo over MoCo v2 (0.4 \vs 0.3 AP$^\text{mask}$)
and is better than ReSim (0.4 \vs 0 AP$^\text{mask}$).
This further verifies the effectiveness of DenseSiam.
Notably, DenseSiam outperforms ReSim, DetCo, DenseCL, and PixPro by 0.7, 0.4, 0.4, and 0.2 AP$^\text{mask}$, respectively.
The results in COCO object detection is consistent with that in instance segmentation.

\myparagraph{Cityscapes Semantic Segmentation.}
We compare DenseSiam with previous methods on semantic segmentation on Cityscapes dataset~\cite{cityscapes} and report the mean IoU (mIoU) of fine-tuned models.
DenseSiam surpasses SimSiam by 0.7 mIoU and
outperforms previous dense representation learning methods ReSim, DetCo, DenseCL, and PixPro by 0.2, 0.5, 1.3, and 0.7 mIoU, respectively.
Notably, image-level unsupervised method such as SimCLR~\cite{simclr}, BYOL~\cite{byol} and SimSiam~\cite{simsiam} show considerable performance gap against 
dense unsupervised methods including DenseSiam, DenseCL, and ReSim, indicating that pre-training with image-level similarity learning is sub-optimal for dense prediction tasks.

\myparagraph{PASCAL VOC Object Detection.}
We further compare DenseSiam with previous methods on PASCAL VOC~\cite{voc} object detection.
We report the original metric AP$_\text{50}$ (AP calculated with IoU threshold 0.5) of VOC and further report COCO-style AP~\cite{wu2019detectron2} and AP$_\text{75}$,
which are stricter criteria in evaluating the detection performance.
DenseSiam show a large improvement of 1.8 AP in comparison with SimSiam~\cite{simsiam}.
Notably, DenseSiam exhibits considerable improvements on AP$_\text{75}$ than AP$_\text{50}$,
suggesting the effectiveness of DenseSiam in learning accurate spatial information.
Its improvement over SimSiam is more than that of DetCo~\cite{detco} (0.8 AP) over MoCo v2, and is on-par with those of DenseCL~\cite{densecl} and ReSim~\cite{resim} over MoCo v2~\cite{mocov2}.

DenseCL, ReSim, and PixPro are 0.2, 0.2, and 1.0 AP better than DenseSiam, respectively.
This phenomenon contradicts the results in the benchmarks of COCO dataset,
although their improvements over their image-level counterparts are consistent across benchmarks.
We hypothesize that the different backbones used in the two benchmarks leads to this phenomenon, where ResNet-50-C4 backbone~\cite{ren2015faster} is used on PASCAL VOC but ResNet-50-FPN~\cite{fpn} is used on COCO.
The hypothesis also explains the inferior performance of DetCo~\cite{detco}:
DetCo conducts contrastive learning on pyramid features but only one feature scale is used when fine-tuning Faster R-CNN on PASCAL VOC dataset.

\subsection{Unsupervised Segmentation Results}
In Table~\ref{tab:unsupervised_sem_seg} we compare DenseSiam with previous state-of-the-art methods in unsupervised semantic segmentation.
DenseSiam achieves new state-of-the-art performance of 16.4 mIoU, surpassing PiCIE by 2 mIoU over all classes.
Imbalanced performance is observed in previous methods between thing and stuff classes,
\eg, PiCIE~\cite{picie} and IIC~\cite{iic} surpass Modified DeepClustering~\cite{deepcluster} on thing classes but fall behind on stuff classes.
In contrast, DenseSiam consistently outperforms previous best results on both thing and stuff classes by more than 2 mIoU.
The results reveal that clustering is unnecessary, whereas clustering is indispensable in previous unsupervised segmentation methods~\cite{picie, iic}.

\begin{table}[t]
  \centering
  \caption{\textbf{Unsupervised Semantic Segmentation}.
  The model is trained and tested on the curated COCO dataset~\cite{iic, picie}.
  ` + aux.' denotes PiCIE or DenseSiam is trained with an auxiliary head
  }\label{tab:unsupervised_sem_seg}
      \tablestyle{3pt}{1.1}
      \scalebox{0.8}{\begin{tabular}{l|x{26}|x{30}x{30}|c}
      \shline
      Method &  mIoU & mIoU\stuff & mIoU\things& Time (h) \\
      \shline
      Modified Deep Clustering~\cite{deepcluster}& 9.8  &22.2  &11.6 & -\\
      IIC~\cite{iic}                & 6.7  &12.0  &13.6 & -\\
      PiCIE~\cite{picie} + aux.     & 14.4 &17.3  &23.8 & 18\\
      \ours{DenseSiam + aux.}& \ours{\bd{16.4}}&\ours{\bd{24.5}}&\ours{\bd{29.5}}&\ours{5} \\
      \shline
      \end{tabular}}
\end{table}

We also calculate the training costs of PiCIE and DenseSiam by measuring the GPU hours used to train the model with similar batch size and backbone.
Because PiCIE needs clustering to obtain pseudo labels before each training epoch, its training time is comprised of the time of label clustering, data loading, and model's forward and backward passes.
In contrast, DenseSiam does not rely on clustering.
In the setting of single GPU training, DenseSiam saves $\sim$72\% training costs in comparison with PiCIE.

\subsection{Ablation Study}
\myparagraph{Unsupervised Pre-training.} We ablate the key components in DenseSiam as shown in Table~\ref{table:ablation_repr}.
The baselines in Table 4a-e are SimSiam (53.5 AP).

\noindent
\emph{i) Effective grid number $K$ in PixSim}: We study the effective number $K$ used in grid sampling in PixSim in Table~\ref{tab:ablation:num_grids}.
The greater the number, the more feature grids will be sampled from the intersected regions and will be used for similarity learning.
With zero grid number DenseSiam degenerates to the SimSiam baseline where no pixel similarity learning is performed.
The results in Table~\ref{tab:ablation:num_grids} shows that 7$\times$7 feature grids is sufficient to improve the per-pixel consistency of dense representation.
We also compare the training memory used in PixSim.
The comparison shows that PixSim only brings 0.1\% $\sim$ 0.3\% extra memory cost in comparison with SimSiam.

\noindent
\emph{ii) Loss weight of PixSim}: We further study the loss weight $\lambda_1$ of $\mathcal{L}_{dense}$ used for per-pixel similarity learning.
The comparative results in Table~\ref{tab:ablation:pixsim_loss_weight} show that with $\lambda_1=1$ we achieve the best performance,
which is equal to the loss weight of the image-level similarity learning.

\noindent
\emph{iii) Loss weight of RegionSim}: We study the loss weight $\lambda_2$ of RegionSim as shown in Table~\ref{tab:ablation:contrast_loss_weight}.
We find that $0.1$ works best and large value of $\lambda_2$ leads decreased performance.

\noindent
\emph{iv) Orders of grid sample, projector, and predictor}: As DenseSiam consists of encoder, projector, and predictor, we study the optimal position where the grid sampler should be introduced.
The results in Table~\ref{tab:ablation:order} show that it is necessary to put the grid sample module after the projector, but the predictor can perform equally well when it is before or behind the grid sample module.

\begin{table}[t]
  \centering
  \caption{\textbf{Ablation studies in unsupervised pre-trainings}.
  All unsupervised representations are based on 800-epoch pre-training on COCO \texttt{train2017}.
  The representations are fine-tuned with Faster R-CNN~\cite{ren2015faster} (C4-backbone) in VOC 2007 trainval + 2012 train and evaluated on VOC 2007 test.
  `Mem.' indicates memory cost measured by Gigabyte (GB).
  The results of fine-tuning are averaged over 5 trials.
  Best settings are bolded and used as default settings
  }\label{table:ablation_repr}
  \centering
  \begin{minipage}[t]{.32\linewidth}
    \centering
    \subcaption{\small{The effective number $K$ in PixSim}}\label{tab:ablation:num_grids}
   \tablestyle{1pt}{1.2}
    \scalebox{0.7}{\begin{tabular}{ c | x{23}x{23}x{23}|x{23}}\shline
      Gird Number & AP & AP$_\text{50}$ & AP$_\text{75}$ & Mem. \\
      \shline
      0  &53.5  &79.7  &59.3&8.06\\
      \hline
      1  &28.8&	53.8&	27.1 & 8.11\\
      3  &53.9&	80.0&	59.4 & 8.12\\
      7  &\bd{54.9}&\bd{80.8}&\bd{60.9} &8.18\\
      9  &54.6&	80.7&	60.6 & 8.21\\
      14 &54.7&	80.6&	60.8 & 8.28\\\shline
      \end{tabular}}
  \end{minipage}\hspace{.02\linewidth}
  \begin{minipage}[t]{.31\linewidth}
    \centering
    \subcaption{\small{The effective loss weight of PixSim}}\label{tab:ablation:pixsim_loss_weight}
  \tablestyle{1pt}{1.2}
    \scalebox{0.7}{\begin{tabular}{ x{23} | x{23}x{23}x{23}}\shline
      $\lambda_1$ & AP & AP$_\text{50}$ & AP$_\text{75}$ \\
      \shline
      0   &53.5  &79.7  &59.3 \\
      \hline
      0.1  &53.3  &79.6 &58.8   \\
      0.3  &53.5  &79.9 &58.8   \\
      0.5  &54.0  &80.2 &60.0   \\
      0.7  &54.0  &79.8 &59.8   \\
      1.0  &\bd{54.9}&\bd{80.8}&\bd{60.9} \\\shline
      \end{tabular}}
  \end{minipage}\hspace{.02\linewidth}
  \begin{minipage}[t]{.31\linewidth}
    \centering
    \subcaption{\small{The effective loss weight of RegionSim}}\label{tab:ablation:contrast_loss_weight}
  \tablestyle{1pt}{1.2}
    \scalebox{0.7}{\begin{tabular}{ x{23} | x{23}x{23}x{23}}\shline
      $\lambda_2$ & AP & AP$_\text{50}$ & AP$_\text{75}$ \\
      \shline
      0     &53.5     &79.7     &59.3 \\\hline
      0.01  &55.3	    &81.0	    &61.3 \\
      0.05  &55.0	    &80.9	    &60.6 \\
      0.1   &\bd{55.5}&\bd{81.1}&\bd{61.5} \\
      0.2   &55.3	    &81.0	    &61.0 \\
      0.5   &55.3	    &80.9	    &61.1 \\\shline
      \end{tabular}}
  \end{minipage}
  \begin{minipage}[t]{.33\linewidth}
    \centering
    \subcaption{\small{The effective order of grid sample, projector, and predictor in PixSim}}\label{tab:ablation:order}
  \tablestyle{1pt}{1.2}
    \scalebox{0.7}{\begin{tabular}{ c | x{23}x{23}x{23}}\shline
      Order & AP & AP$_\text{50}$ & AP$_\text{75}$ \\
      \shline
      -                         &53.5  &79.7  &59.3 \\\hline
      grid.~{+}~proj.~{+}~pred. &53.4	 &79.6	&59.0 \\
      proj.~{+}~grid.~{+}~pred. &54.7	 &80.6	&60.4 \\
      proj.~{+}~pred.~{+}~grid. &\bd{54.9}&\bd{80.8}&\bd{60.9} \\\shline
      \end{tabular}}
  \end{minipage}\hspace{.01\linewidth}
  \begin{minipage}[t]{.37\linewidth}
    \centering
    \subcaption{
      \small{Regions to be focused in global branch. `in.' indicates the intersected regions}}\label{tab:ablation:region}
  \tablestyle{1pt}{1.2}
    \scalebox{0.7}{\begin{tabular}{ x{50}|x{50}|x{23}x{23}x{23}}\shline
      Image-level & Pixel-level & AP & AP$_\text{50}$ & AP$_\text{75}$ \\
      \shline
      global & N/A &53.5 &79.7 &59.3 \\
      in. & N/A &0.0&	0.0	&0.0 \\
      \hline
      global & in.&\bd{54.9}&\bd{80.8}&\bd{60.9} \\
      in. & in. & 35.6	&61.5	&35.6 \\\shline
      \end{tabular}}
  \end{minipage}
  \hspace{.01\linewidth}
  \begin{minipage}[t]{.25\linewidth}
    \centering
    \subcaption{\small{Suitable start epoch of RegionSim}}\label{tab:ablation:start_epoch}
  \tablestyle{1pt}{1.2}
    \scalebox{0.7}{\begin{tabular}{ x{48} | x{23}x{23}x{23}}\shline
      Start epoch & AP & AP$_\text{50}$ & AP$_\text{75}$ \\
      \shline
      never & 54.9&80.8&60.9 \\ \hline
      0.4  &37.7	&64.8	&38.4\\
      0.5  &\bd{55.5}&\bd{81.1}&\bd{61.5} \\
      0.6  &55.0	&80.8	&60.4 \\\hline
      w/o PixSim & 52.8 &79.2 &58.0 \\ \shline
      \end{tabular}}
  \end{minipage}
\end{table}

\noindent
\emph{v) Regions to focus in global branch}: We also study the regions that should be focused in global branch in Table~\ref{tab:ablation:region}, as an image-level similarity learning branch (SimSiam) is kept to facilitate training.
The abbreviations `global' and `in.' indicate whether the similarity learning is conducted with the whole image or the intersected regions between two views.
To maximize per-pixel consistency, PixSim is always conducted with the intersected regions.
We find that the focusing on the whole image when conducting image-level similarity learning is always important in both SimSiam and DenseSiam.
Only using the intersected regions will lead to feature `collapsing' (second row with zero accuracy in downstream tasks) in SimSiam since it makes some learning shortcuts for the model.

\noindent
\emph{vi) Start point of RegionSim}: The start point of RegionSim based on PixSim (54.9) matters as shown in Table~\ref{tab:ablation:start_epoch}. 
The label prediction quality is not accurate and may lead to a wrong optimization direction if RegionSim is applied at a wrong time.
Starting RegionSim at the middle point of training yields the best performance.
We further try only using RegionSim without PixSim (last row in Table~\ref{tab:ablation:start_epoch}).
Only adding RegionSim degrades fine-tuning results on by 2.7 AP,
this also implies that RegionSim needs PixSim to produce meaningful groups.

\myparagraph{Unsupervised Semantic Segmentation.}
We also study the effectiveness of the components in DenseSiam for unsupervised semantic segmentation.
Directly applying PixSim without any further modification yields 10.1 mIoU, which already surpasses many previous methods~\cite{deepcluster, iic} (9.8 and 6.7 mIoU).
Adding CE loss further improves the performance by making the feature space more compact and discriminative.
After adding the auxiliary head, the model already surpasses the previous state-of-the-art method PiCIE.
RegionSim further brings 1.4 mIoU of improvement.
 
\begin{table}[t]
    \centering
    \caption{\textbf{Ablation study in unsupervised segmentation}.
    `Aux.' indicates auxiliary head
    }\label{tab:ablation:unsupervised_sem_seg}
    \tablestyle{3pt}{1.1}
    \scalebox{0.75}{\begin{tabular}{x{28}|x{24}|x{22}|x{28}|x{26}|x{28}x{28}}\shline
      PixSim & Aux. & CE & Region. & mIoU & mIoU\stuff & mIoU\things \\ \shline
      \checkmark &&&&10.1 &19.0&17.7 \\
      \checkmark & \checkmark & &&11.1& 20.4 & 22.3 \\
      \checkmark & \checkmark &\checkmark &&15.0&24.8&23.4 \\
      \checkmark & \checkmark & \checkmark &\checkmark&16.4&24.5&29.5 \\\shline
    \end{tabular}}
\end{table}

\section{Conclusion}
Different dense prediction tasks essentially shares the similar goal of optimizing the spatial consistency between views of images.
DenseSiam exploits such a property and unanimously solves unsupervised dense representation learning and unsupervised semantic segmentation within a Siamese architecture. 
DenseSiam optimizes similarity between dense predictions at pixel level by PixSim and at region level by RegionSim,
with neither negative pixel pairs, momentum encoder, manual region crops, nor heuristic masks,
which are \emph{all unnecessary} as revealed by DenseSiam to obtain a strong dense representation for downstream tasks.
Its unsupervised semantic segmentation performance also achieves the new state-of-the-art.

\myparagraph{Acknowledgements.}
This study is supported under the RIE2020 Industry Alignment Fund Industry Collaboration Projects (IAF-ICP) Funding Initiative, as well as cash and in-kind contribution from the industry partner(s).
The work is also suported by Singapore MOE AcRF Tier 2 (MOE-T2EP20120-0001) and NTU NAP Grant.
Jiangmiao Pang and Kai Chen are partially supported by the Shanghai Committee of Science and Technology, China (Grant No. 20DZ1100800).

\appendix
\setcounter{table}{0} 
\setcounter{figure}{0}
\setcounter{equation}{0}
\renewcommand{\thetable}{A\arabic{table}}
\renewcommand\thefigure{A\arabic{figure}} 
\renewcommand\theequation{A\arabic{equation}}

\section{Implementation Details}
\myparagraph{Data Augmentation.} We use the same data augmentation techniques as those used in previous methods for a fair comparison.

\noindent
\emph{i) Unsupervised pre-training}:
We use existing augmentation modules in PyTorch~\cite{pytorch} and describe them using the same notations as the following.
Specifically, for geometric augmentation, we use \texttt{RandomResizedCrop} with scale in [0.2,1.0] and \texttt{RandomHorizontalFlip}.
For color augmentations we use \texttt{ColorJitter} and \texttt{RandomGrayscale} with probabilities of 0.8 and 0.2, respectively.
The jittering strength of brightness, contrast, saturation, and hue are 0.4, 0.4, 0.4, and 0.1 in \texttt{ColorJitter}, respectively.
Blurring augmentation~\cite{simclr} is also applied using a Gaussian kernel with std in [0.1, 2.0].
These hyper-parameters are the same as those adopted in previous methods~\cite{simsiam, mocov2, densecl, byol, simclr}.

\noindent
\emph{i) Unsupervised semantic segmentation}:
We apply the same augmentations as those used in PiCIE~\cite{picie} in unsupervised semantic segmentation for a fair comparison.
The augmentations include color jittering, gray scale, blurring, cropping, and flipping.
The color jittering augmentations consists of jittering brightness, contrast, saturation, and hue.
These jittering transformations are randomly applied with probabilities of 0.8 with strength of 0.3, 0.3, 0.3, and 0.1, respectively.
The gray scale augmentation is randomly applied with a probability of 0.2.
Random crop is used with scale in [0.5,1.0].
Different from those augmentations used in unsupervised pre-training,
the augmentations are sampled first and replayed with similar parameters in each epoch following PiCIE~\cite{picie}.

\myparagraph{Transfer Learning Results of PixPro.}
In Table~\ref{tab:transfer}, we compare DenseSiam with PixPro~\cite{pixpro}.
Since PixPro~\cite{pixpro} only reports the transfer learning results of 100-epoch or 400-epoch pre-training in ImageNet (IN1k),
we conduct 200-epoch pre-training using the official code~\footnote{\url{https://github.com/zdaxie/PixPro}} for a fair comparison.
The pre-trained backbone are then fine-tuned using the same evaluation protocol of transfer learning as described in Sec.~\ref{sec:exps:settings}.
  
\myparagraph{Exploitation of Correspondence.}
Previous methods explore different strategies to build and optimize dense correspondence between two views.
DenseCL~\cite{densecl} compares feature similarity to build the correspondence between pixels.
PixPro~\cite{pixpro} connects pixels by the distance of their coordinates in the original image.
There are also attempts~\cite{resim, scrl} to use manual crops and maximize the similarity between similar crops in different views.
DenseSiam uses the location and regions for similarity learning of different granularities.

For a fair comparison, we study these strategies under the same architecture of PixSim with a switch between strategies.
Specifically, we keep the same architecture and symmetrized loss of PixSim and implement these strategies strictly following their official code releases.
To study feature similarity used in DenseCL~\cite{densecl}, we calculate feature similarities between pixels and link the most similar pixels between views.
To study pixel distance used in PixPro~\cite{pixpro},  we calculate the coordinate distances of pixels and link the pixels having the close locations.
To study candidate regions used in ReSim~\cite{resim}, we use anchors in the intersected regions generated by sliding windows.

Table~\ref{tab:analysis:correspondence} show that using feature similarity~\cite{densecl} significantly decreases the performance.
This is because PixSim does not use negative pixel pairs, a prerequisite for the strategy to work.
Using pixel coordinates only brings marginal improvements.
Using location correspondence in PixSim is much simpler and more effective than candidate regions.
DenseSiam yields the best results by exploiting the correspondence built with both location and region embeddings.

\begin{table}[t]
    \centering
    \begin{minipage}[t]{0.47\linewidth}
      \caption{\textbf{Analysis:} Strategies for building visual correspondence
      }\label{tab:analysis:correspondence}
      \tablestyle{3pt}{1.1}
      \scalebox{0.7}{\begin{tabular}{l |x{22}|x{24}x{24}}\shline
      Strategies & AP & AP$_\text{50}$ & AP$_\text{75}$ \\\shline
      SimSiam                             &53.5 &79.7 &59.3\\\hline
      feature similarity~\cite{densecl}   &36.3 &63.5 &36.1\\
      pixel distance~\cite{pixpro}        &53.7	&79.3 &58.5\\
      candidate regions~\cite{resim, scrl}&54.7 &79.9 &60.5\\\hline
      location                    &54.9 &80.8 &60.9\\
      location + region embeddings &\bd{55.5} &\bd{81.1} &\bd{61.5}\\\shline
      \end{tabular}}
    \end{minipage}
    \begin{minipage}[t]{0.49\linewidth}
      \caption{\textbf{Analysis:} Strategies for grid sampling
      }\label{tab:analysis:sample_strategies}
      \tablestyle{3pt}{1.1}
      \scalebox{0.8}{
          \begin{tabular}{l |c|c|c}\shline
              Strategies & AP & AP$_\text{50}$ & AP$_\text{75}$ \\\shline
              uniform (7$\times$7 regular grid)  &54.9 &80.8 &60.9\\\hline
              uniform ($k=1, \beta=0.0$)   &54.7 &80.7 &60.3\\
              midly biased ($k=3, \beta=0.75$)   &55.0&80.7&60.8\\
              heavily biased ($k=10, \beta=1.0$)   &54.3&80.3&60.0\\\shline
          \end{tabular}}
    \end{minipage}
\end{table}

\begin{figure}[t]
    \centering
    \caption{\small{
		\textbf{Visualization} of pseudo categories of pixels produced by PixSim in unsupervised representation learning.
    }}
    \vspace{-9pt}
    \begin{minipage}{0.49\textwidth}
        \includegraphics[width=0.49\textwidth]{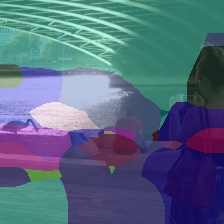}
        \includegraphics[width=0.49\textwidth]{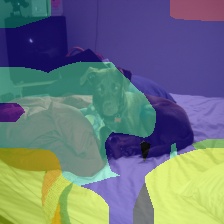}
    \end{minipage}
    \begin{minipage}{0.49\textwidth}
        \includegraphics[width=0.49\textwidth]{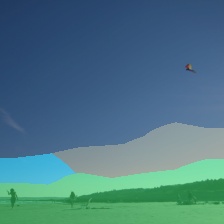}
        \includegraphics[width=0.49\textwidth]{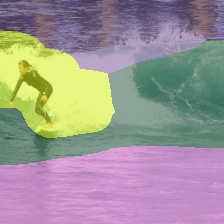}
    \end{minipage}
    \vspace{-12pt}
\end{figure}

\section{Analysis}

\myparagraph{Grid Sampling Strategies.}
We tried a hard example mining strategy proposed in PointRend.
Specifically, it first over-generates candidate points by randomly sampling $kN$ points ($k>1$) from a uniform distribution.
Then it estimates the similarities by Eq.3 between the embeddings of these points from both views.
Finally, it selects the most dissimilar $\beta N (\beta \in [0, 1])$ points from the $kN$ candidates and
sample the remaining $(1-\beta)N$ points from a uniform distribution.

The results in Table~\ref{tab:analysis:sample_strategies} shows that the strategy marginally improves the performance
and only training on hard examples degrades the results.
More strategies can be explored in future research.

\myparagraph{Visualization of masks.}
When using cross-entropy similarity, $\texttt{softmax}(z'_1)$ can be treated as a segmentation map,
in which the pseudo categories of each pixel can be obtained by \texttt{argmax} over the channel dimension~\cite{simsiam}.
We visualize the pixels' pseudo categories by different colors in the figures below.
The results show that the pixels are grouped into different pseudo categories without supervision,
and the features gathered by these masks thus contain region-level information,
which are then forced to have consistency across views in RegionSim.

\clearpage
%
%
\bibliographystyle{splncs04}
\bibliography{sections/mainbib}
\end{document}